\documentclass[lettersize,journal]{IEEEtran}

\usepackage{times}
\usepackage{graphics,booktabs,epsfig,enumerate}
\usepackage{amsfonts,amssymb,amsmath,amsbsy,bm,paralist,theorem,ifthen,color}
\usepackage[mathscr]{euscript} 
\usepackage{enumerate}
\usepackage{amsmath}
\usepackage{longtable}
\usepackage{rotating}
\usepackage{graphicx}
\usepackage{diagbox}
\usepackage{CJK}
\usepackage{multirow}
\usepackage{float}
\usepackage{stfloats}
\usepackage{booktabs}
\usepackage{longtable}
\usepackage{rotating}
\usepackage{enumitem}
\usepackage{subfigure}
\usepackage[justification=raggedright]{caption}
\usepackage{caption}
\newtheorem{lemma}{Lemma}

\usepackage{braket}
\usepackage{algorithm}  
\usepackage{algorithmic}

\begin{document}

\title{Study of Robust Direction Finding Based on Joint Sparse Representation }

\author{Yingsong Li,~\IEEEmembership{Senior Member,~IEEE,} Wudang Xiao, Luyu Zhao,~\IEEEmembership{Senior Member,~IEEE}, Zhixiang Huang,~\IEEEmembership{Senior Member,~IEEE}, Qiang Li, Liping Li, Rodrigo C. de Lamare,~\IEEEmembership{Senior Member,~IEEE}

\thanks{Y. Li, W. Xiao, L. Zhao, Q. Li, L. Li, Z. Huang are with the Key Laboratory of Intelligent Computing and Signal Processing Ministry of Education, Anhui University, Hefei, Anhui 230601, China. R. C. de Lamare is with PUC-Rio, Brazil, and with the University of York, UK. (e-mail:liyingsong@ieee.org).}

\thanks{Rodrigo. C. de Lamare is with the Centre for Telecommunications Research (CETUC), Pontifical Catholic University of Rio de Janeiro (PUC-Rio), G$\mathbf{\acute{a}}$vea, 22451-900, Brazil, and the Department of Electronic Engineering, University of York, York, YO10 5DD, UK (e-mail: delamare@cetuc.puc-rio.br).}
}

\linespread{0.95}
\markboth{}%
{Shell \MakeLowercase{\textit{et.al.}}: Bare Demo of IEEEtran.cls for IEEE Journals}

\maketitle

\begin{abstract}
Standard Direction of Arrival (DOA) estimation methods are typically derived based on the Gaussian noise assumption, making them highly sensitive to outliers. Therefore, in the presence of impulsive noise, the performance of these methods may significantly deteriorate. In this paper, we model impulsive noise as Gaussian noise mixed with sparse outliers. By exploiting their statistical differences, we propose a novel DOA estimation method based on sparse signal recovery (SSR). Furthermore, to address the issue of grid mismatch, we utilize an alternating optimization approach that relies on the estimated outlier matrix and the on-grid DOA estimates to obtain the off-grid DOA estimates. Simulation results demonstrate that the proposed method exhibits robustness against large outliers.
\end{abstract}

\begin{IEEEkeywords}
DOA estimation, outliers, sparse signal recovery. \vspace{-0.45em}
\end{IEEEkeywords}

\IEEEpeerreviewmaketitle

\vspace{-3mm}

\section{Introduction}

\IEEEPARstart Direction of Arrival (DOA) estimation stands as one of the pivotal subjects in array signal processing, playing a key role in fields such as radar, sonar and wireless communications \cite{krim1996two}. Over the past few decades, DOA estimation techniques have been introduced and extensively studied in the academic literature, such as classical subspace-based approaches (e.g.,MUSIC,ESPRIT) \cite{schmidt1986multiple,jio_doa,alr_doa,roy1989esprit,mskaesprit}, which rely on the eigenvalue decomposition of the sample covariance matrix. However, they require a substantial number of snapshots to obtain accurate estimates and their performance degrades significantly in the presence of non-Gaussian noise, such as impulsive noise. Variants of subspace-based methods have been proposed to deal with impulsive noise, such as ROC-MUSIC \cite{tsakalides1996robust}, FLOM-MUSIC \cite{liu2001subspace}. These methods utilize fractional lower-order statistics instead of second-order statistics to tackle impulsive noise, but still require a significant number of samples. Inspired by robust statistics, the $l_p$-MUSIC method has been introduced \cite{zeng2013ell}, replacing the traditional Frobenius norm with the $l_p$-norm (1 $\leq$ $p$ $<$ 2) to minimize residuals, which is less sensitive to outliers than the former. Based on the work of \cite{zeng2013ell}, \cite{ma2020generalised} introduced a generalized maximum complex correntropy criterion for suppressing outliers in impulse noise.

Recently, the development of compressive sensing theory has led to the emergence of numerous DOA estimation methods based on sparse representation \cite{yang2021convex,leite2021}. These methods encompass greedy algorithms \cite{tropp2007signal}, convex optimization \cite{liu2017off}, and sparse Bayesian learning \cite{yang2012off}. Unlike subspace-based methods, sparse signal recovery methods \cite{jidf,sjidf,dce} based on compressive sensing theory offer advantages in low signal-to-noise ratios, fewer snapshots, and highly correlated sources. In practice, the DOAs may not be aligned precisely with the predefined grid, leading to certain angle deviations. To address this issue, two methods are commonly employed. One approach involves increasing the number of predefined grid points; however, as the grid density increases, the correlation between adjacent atoms in the dictionary grows significantly. This contradicts the Restricted Isometry Property (RIP) condition of compressive sensing and comes with a high computational cost \cite{liu2022adaptive}. The other approach is to adopt an off-grid strategy, where the grid spacing is treated as a parameter. It is estimated by first or second-order Taylor expansions, effectively approximating it \cite{zhang2019off,huang2022off}. In these approaches, employing the $l_2$-norm as the residual term is inefficient, as it is equally sensitive to impulse noise, making it less robust in the presence of outliers.

Robust sparse recovery methods have been proposed to mitigate the impact of outliers in \cite{suzuki2020robust,wen2016robust}. Different norms are used in \cite{suzuki2020robust,wen2016robust} as residual fitting terms, which are less sensitive to large outliers as compared to least-squares. This can, to some extent, reduce the influence of impulsive noise or outliers on DOA estimation. In \cite{suzuki2023sparse}, a novel sparse robust signal recovery framework was introduced. This method exploits the sparsity of outliers and the statistical properties of Gaussian noise, making it highly robust against outliers. In \cite{dai2017sparse}, a sparse Bayesian learning algorithm for DOA estimation under impulsive noise conditions was introduced, which exhibits high resolution and accuracy.

In this work, we introduce a novel DOA estimation technique for impulsive noise scenarios, where the impulsive noise is modeled as a combination of Gaussian noise and outliers and an objective function is constructed to exploit the sparsity of outliers and the statistical properties of Gaussian noises. This objective function incorporates the Minimax Logarithmic Concave (MLC) Function \cite{zhang2023tensor} as a sparsity-inducing term for outliers. The proposed design significantly enhances the algorithm's robustness, particularly when dealing with severe impulsive noise conditions. Furthermore, we extend this approach to Multiple Measurement Vectors (MMV) and introduce an alternating optimization algorithm to tackle the grid mismatch problem, which employs rough estimates of the on-grid angles and the outlier matrix to obtain off-grid DOA estimates. 
\vspace{-3mm}
\section{Signal Model}

Consider $K$ far-field narrowband sources $s_k(t), (k=1,2,...,K)$ impinging on a uniform linear array (ULA) with $M$ omnidirectional elements from directions of $\boldsymbol{\theta}=[\theta_{1},...,\theta_{K}]^T$. The array output at snapshot $t$ is given by
\begin{equation}
\begin{aligned}
\mathbf{y}(t)&=\sum_{k=1}^{K}{\mathbf{a}(\theta_k)s_{k}(t)}+\mathbf{n}(t) \\
&=\mathbf{A}(\boldsymbol{\theta})\mathbf{s}(t)+\mathbf{n}(t),
\end{aligned} \label{model}
\end{equation}
where $\mathbf{y}(t)=[y_1(t),...,y_M(t)]^T$ is the array output, $\mathbf{A}(\boldsymbol{\theta})=[\mathbf{a}(\theta_1),...,\mathbf{a}(\theta_K)]$ is the array manifold matrix, $\mathbf{a}(\theta_k)=[1,...,e^{-j2\pi (M-1)d/ \lambda_s\sin({\theta_k})}]^T$ is the time delay of the $k$th source at each array element, $\mathbf{s}(t)=[s_1(t),...,s_K(t)]^T$ are the source waveforms, $\lambda_s$ and $d$ denote the wavelength of sources and the distance between adjacent array elements, respectively, $d=\lambda_s /2$, and $\mathbf{n}(t)$ is the unknown noise. The model in \eqref{model} after $T$ snapshots can be rewritten in matrix form as
\begin{equation}
\mathbf{Y}=\mathbf{A(\boldsymbol{\theta})S}+\mathbf{N},
\label{model2}
\end{equation}
where $\mathbf{S}=[\mathbf{s}(1),...,\mathbf{s}(T)]$ and $\mathbf{N}=[\mathbf{n}(1),...,\mathbf{n}(T)]$.

Given $\mathbf{A}$ and $\mathbf{Y}$, the DOA estimation problem can be transformed into a sparse signal recovery problem. Exploiting the sparsity of the number of sources in the spatial domain, the angular space $[-\pi/2, \pi/2]$ is divided into $N$ equidistant grids $\overline{\boldsymbol{\theta}}=[\overline\theta_1,...,\overline\theta_N]^T$, satisfying the condition $N \gg K$ . The sparse representation of \eqref{model2} can be rewritten as
\begin{equation} \mathbf{Y}=\mathbf{A(\overline{\boldsymbol{\theta}})\mathbf{X}}+\mathbf{N},
\end{equation}
where $\mathbf{A(\overline{\boldsymbol{\theta}})}=[\mathbf{a}(\overline{\theta}_1),...,\mathbf{a}(\overline{\theta}_N)]$ is an overcomplete dictionary  and $\mathbf{X}=[\mathbf{x}(1),...,\mathbf{x}(T)] \in \mathbf{C}^{N \times T}$ is a row sparse matrix with $K$ nonzero rows. The indices of rows in $\mathbf{X}$ correspond to the positions of DOAs on the grid. If the $n$th row of $\mathbf{X}$ is nonzero, it represents the presence of a source in direction $ \overline\theta_n$. Therefore, the objective of DOA estimation is to compute $\mathbf{X}$.


\section{Proposed DOA Estimation Algorithm}

In this section, we detail the proposed DOA estimation algorithm based on the MLC function and an on-grid model. 

\subsection{Minimax Logarithmic Concave Function}
We introduce a novel non-convex MLC function $\phi:\mathbb{C} \rightarrow \mathbb{R}^{+}$ given by
\begin{equation}
        \phi(\textit{x})=\left\{
        \begin{aligned}
            & \lambda\log{(\frac{\lvert \textit{x} \rvert}{\eta}+1)}-\frac{\log^2{(\frac{\lvert \textit{x} \rvert}{\eta}+1)}}{2\gamma}, \lvert \textit{x} \rvert \leq \eta e^{\gamma \lambda} - \eta \\
            & \frac{\gamma \lambda^2}{2}, \quad \quad \quad  \lvert \textit{x} \rvert > \eta e^{\gamma \lambda} - \eta
        \end{aligned}
        \right.,
\end{equation}
where $\lambda >0$, $\gamma >0$, $\eta >0$. Compared to functions such as $l_1$, $l_p$ and MC \cite{suzuki2020robust}, the MLC function exhibits better sparsity-inducing properties than other approaches, as illustrated in Fig. 1.
\begin{figure}[ht]

\centerline{\includegraphics[width=8cm,height=8cm]{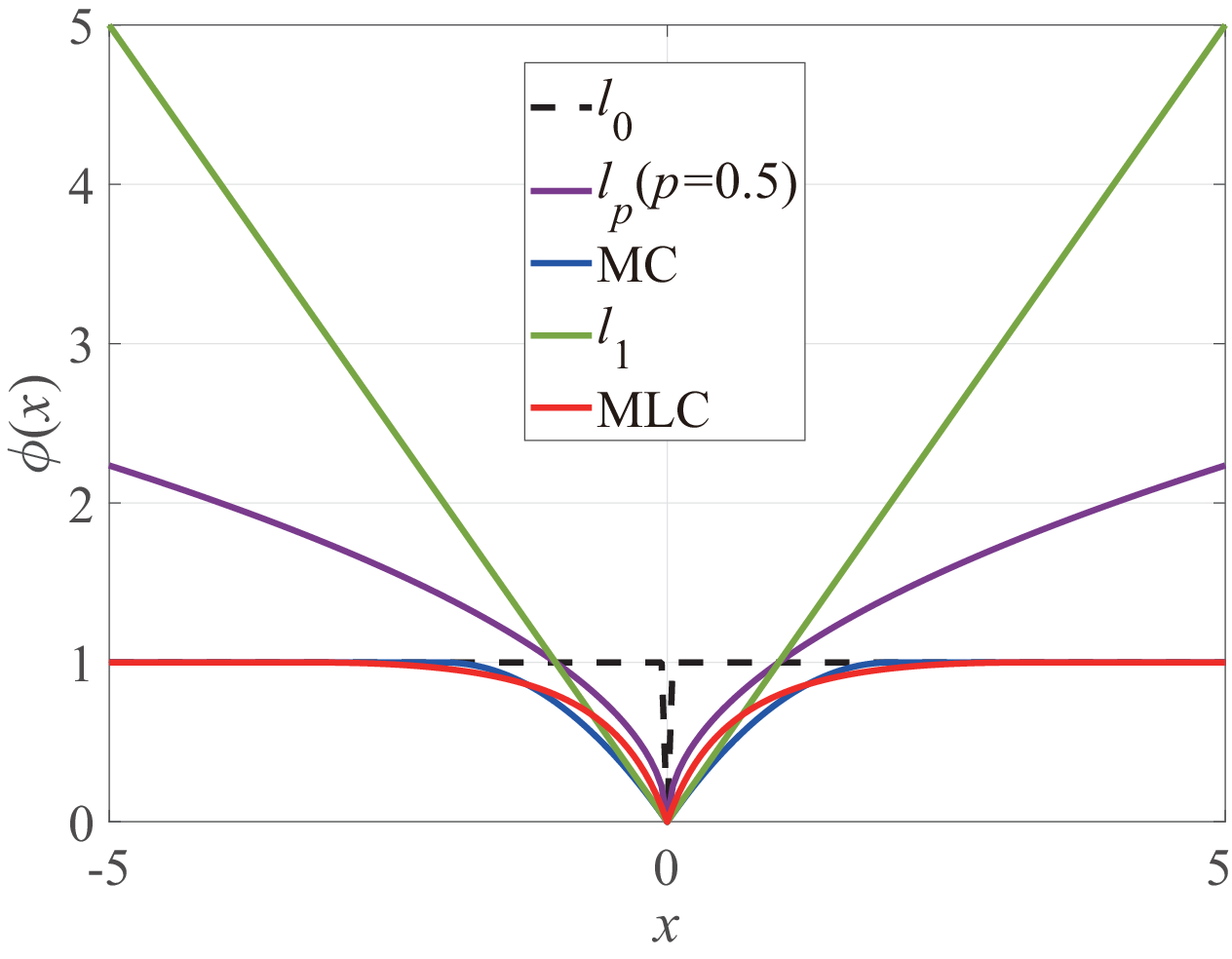}}
    \caption{Sparsity-inducing functions. Here $\lambda=1$, $\gamma=2$, $\eta=0.4$ for MLC; $\lambda=1$, $\gamma=2$ for MC.}
    \label{sparse functions}
\end{figure}

Before introducing the proposed algorithm, we introduce two highly useful lemmas.
\begin{lemma}
    For $\lambda>0$, $\gamma>0$, $\eta>0$, the function $\phi(\textit{x}):\mathbb{C} \rightarrow \mathbb{R}^{+}$ is equivalent to the optimal solution of the following optimization problem:
    \begin{equation}
        \phi(\textit{x})=\min\limits_{w \in \mathbb{R}^+} w\log{(\frac{\lvert \textit{x} \rvert}{\eta}+1)} + \frac{\gamma}{2}(w-\lambda)^2,
    \end{equation}
    where
    \begin{equation}
        w^\ast = \max\left\{\lambda-\frac{\log{(\frac{\lvert \textit{x} \rvert}{\eta}+1)}}{\gamma},0\right\}.
    \end{equation}
\end{lemma}

Please refer to \cite[Theorem 2]{zhang2023tensor} for a detailed proof of this lemma. While the proof is based on real values, it remains valid for complex values.

\begin{lemma}
    Let $\eta >0$, $\mu >0$, $\textit{x}$, $c \in \mathbb{C}$, the proximity operator of function $\log(\lvert \textit{x} \lvert + \eta)$ can be expressed as
    \begin{equation}
        \begin{array}{cc}
             {\kern -30pt} {\rm{prox}}_{\mu,\log(\cdot)}(c)=\arg\min\limits_{\textit{x}} \mu\log(\lvert \textit{x} \lvert + \eta) + \frac{1}{2}\lvert \textit{x}-c\lvert^2 \\
             {\kern -10pt}= \left\{
            \begin{aligned}
                 & 0, \quad  {\rm{if}} \ \lvert c \lvert \leq 2\sqrt{\mu} - \eta \\
                 & \alpha, \quad  {\rm{if}} \  \lvert c \lvert > 2\sqrt{\mu} - \eta
            \end{aligned} 
            \right.\quad
        \end{array},
    \end{equation}
    where
    \begin{equation}
        \begin{array}{cc}
             \alpha = \frac{-\eta+\lvert c \lvert+\sqrt{(\lvert c \lvert)^2-4\mu}}{2} \frac{\lvert c \lvert}{c}.
        \end{array}
    \end{equation}
\end{lemma}

\subsection{On-Grid Model}
In the presence of impulsive noise, the probability density function (PDF) of  $\mathbf{N}$ exhibits heavier tails compared to the Gaussian distribution, leading to the presence of a few outliers. Therefore, the data model (3) can be rewritten as
\begin{equation}
\mathbf{Y}=\mathbf{A(\overline{\boldsymbol{\theta}})\mathbf{X}}+\mathbf{W} + \mathbf{O},
\end{equation}
where $\mathbf{W}$ is Gaussian noise, while $\mathbf{O}$ represents dot-sparse outliers, which do not exhibit row-sparsity characteristics.

In most literature on robust signal recovery, the loss functions $F(\mathbf{Y-AX})$ fail to adequately distinguish between the statistical properties of Gaussian noise and outliers. For example, the Huber loss function is insensitive to Gaussian noise for small perturbations, but its impact increases as the error values grow \cite{ollila2015multichannel}. On the other hand, sparsity-inducing loss functions such as MC \cite{suzuki2020robust}, $l_1$, $l_p$) are non-differentiable at the origin, meaning that small perturbations have a significant impact, thereby not fully leveraging the statistical properties of Gaussian noise.

Therefore, to better exploit statistical properties of 
impulsive noise, we propose a robust sparse signal recovery formulation based on MLC and $l_1$. The formulation can simultaneously utilize the dot-sparsity characteristics of outliers $\mathbf{O}$ and the row-sparsity characteristics of the signal $\mathbf{X}$. Specifically, we formulate the following optimization problem:
\begin{equation}
    \begin{array}{cc}
       {\kern -50pt} \min\limits_{\mathbf{X,W,O}} \
        \frac{1}{2} \Vert \mathbf{W} \Vert_{F}^2 + \lambda_1 \Vert \mathbf{X} \Vert_{2,1} + \lambda_2 \phi(\mathbf{O}) 
        \\ \mathrm{s.t.}~\mathbf{AX+W+O=Y}
    \end{array},
\end{equation}
where $\phi(\mathbf{O})=\sum_{m=1}^{M}\sum_{t=1}^{T}{\mathbf{O}_{m,t}}, $ $\lambda_1$, $\lambda_2 \in \mathbb{R}^{+}$ is the regularization parameter. The augmented Lagrangian function of (10) can be expressed as
\begin{equation}
\begin{array}{c}
{\kern -20pt}\mathcal{L}_\rho(\mathbf{X,W,O,U}) = \frac{1}{2}\Vert\mathbf{W} \Vert_{F}^2 + \lambda_1 \Vert \mathbf{X} \Vert_{2,1}+\lambda_2 \phi(\mathbf{O}) \\
\\[0.01mm]
-\Braket{\mathbf{U}, \mathbf{AX+W+O-Y}} + \frac{\rho}{2} \Vert \mathbf{AX+W+O-Y} \Vert_F^2
\end{array},
\end{equation}
where $\mathbf{U}\in\mathbb{C}^{M \times T}$ is the dual variable, $\rho \in \mathbb{R}^{+}$ is a penalty parameter. We employ the alternating direction method of multipliers (ADMM) to obtain the following recursions \cite{lu2011fast}:
\begin{equation}
    \begin{array}{cc}
       {\kern -105pt} \mathbf{X}^{l+1} = \arg\min\limits_{\mathbf{X}} \lambda_1 \Vert \mathbf{X} \Vert_{2,1} \\ + \frac{\rho}{2} \left \Vert \mathbf{AX}+\mathbf{W}^l+\mathbf{O}^l-\mathbf{Y}-\frac{\mathbf{U}^l}{\rho} \right \Vert_F^2
    \end{array},
\end{equation}
\begin{equation}
    \begin{array}{cc}
        {\kern -112pt} \mathbf{O}^{l+1} = \arg\min\limits_{\mathbf{O}} \lambda_2\phi(\mathbf{O}) \\ 
        {\kern 10pt} + \frac{\rho}{2} \left \Vert \mathbf{A}\mathbf{X}^{l+1}+\mathbf{W}^l+\mathbf{O} -\mathbf{Y}-\frac{\mathbf{U}^l}{\rho} \right \Vert_F^2
    \end{array},
\end{equation}
\begin{equation}
    \begin{array}{cc}
       {\kern -90pt} \mathbf{W}^{l+1} = \arg\min\limits_{\mathbf{W}} \frac{1}{2} \Vert \mathbf{W} \Vert_{F}^2 \\ 
       {\kern 37pt} + \frac{\rho}{2} \left \Vert \mathbf{A}\mathbf{X}^{l+1}+\mathbf{W}+\mathbf{O}^{l+1} -\mathbf{Y}-\frac{\mathbf{U}^l}{\rho} \right \Vert_F^2
    \end{array},
\end{equation}
\begin{equation}
    \begin{array}{cc}
         \mathbf{U}^{l+1}=\mathbf{U}^l-\rho(\mathbf{A}\mathbf{X}^{l+1}+\mathbf{W}^{l+1} + \mathbf{O}^{l+1}-\mathbf{Y}), 
    \end{array}
\end{equation}
where $\mathbf{X}^l,\mathbf{O}^l,\mathbf{W}^l$ represent the values of the variables at the $l$th iteration. The solution for $\mathbf{X}$ via subproblem (12) is an extension of the $l_1-l_2$ minimization problem under the MMV scenario. Since this problem does not have an exact closed-form solution, we approximate the objective function by the Taylor expansion of its second term. For $\mathbf{X}^l$, we have
\begin{equation}
    \begin{array}{cc}
        {\kern -20pt}\left \Vert \mathbf{AX}+\mathbf{W}^l+\mathbf{O}^l-\mathbf{Y}-\frac{\mathbf{U}^l}{\rho} \right \Vert_F^2= \Vert \mathbf{AX}-\mathbf{V}^l \Vert_F^2  \\ 
        \thickapprox \Vert \mathbf{A}\mathbf{X}^l - \mathbf{V}^l \Vert_F^2 + 2\Braket{\mathbf{X}-\mathbf{X}^l , \mathbf{D}(\mathbf{X}^l)} +\frac{1}{\beta} \Vert \mathbf{X}-\mathbf{X}^l \Vert_F^2
    \end{array},
\end{equation}
where $\beta>0$ is a proximal parameter,  $\mathbf{V}^l=\mathbf{Y}+\frac{\mathbf{U}^l}{\rho}-\mathbf{W}^l-\mathbf{O}^l$, $\mathbf{D}(\mathbf{X}^l)=\mathbf{A}^H(\mathbf{A}\mathbf{X}^l-\mathbf{V}^l)$ is the gradient of the function $\Vert \mathbf{AX}-\mathbf{V}^l \Vert_F^2$ considering $\mathbf{X}^l$. Hence, to solve for $\mathbf{X}$ via subproblem (12) is equivalent to
\begin{equation}
    \begin{array}{cc}
        {\kern -12.6pt}\mathbf{X}^{l+1}  = \arg\min\limits_{\mathbf{X}} \lambda_1 \Vert \mathbf{X} \Vert_{2,1} + \frac{\rho}{2\beta} \Bigg \Vert \mathbf{X} - \mathbf{X}^l + \beta\mathbf{D}(\mathbf{X}^l) \Bigg \Vert_F^2 \\ 
        {\kern -78.5pt} = \text{soft} \left (\mathbf{X}^l-\beta\mathbf{D}(\mathbf{X}^l),\frac{\lambda_1\beta}{\rho} \right)
    \end{array},
\end{equation}
where $\text{soft}(\cdot,\cdot)$ is a soft-thresholding function given by 
\begin{equation}
    \begin{array}{cc}
        {\kern -25pt} \rm{soft}(\mathbf{C},\tau) =\arg\min\limits_{\mathbf{X}} \tau \Vert \mathbf{X} \Vert_{2,1} + \frac{1}{2} \Vert \mathbf{X-C} \Vert_{F}^2 \\
         = \left\{
        \begin{aligned}
             & \frac{\Vert \mathbf{C}(i,:) \Vert_2 - \tau}{\Vert \mathbf{C}(i,:) \Vert_2} \mathbf{C}(i,:), \text{if} \ \Vert \mathbf{C}(i,:) \Vert_2 \geq \tau \\
             & 0, \quad \quad \quad \quad \quad \quad \quad \quad \ \rm otherwise
        \end{aligned} 
        \right.\quad
    \end{array},
\end{equation}
where $\mathbf{C}(i,:)$ is $i$th row of matrix $\mathbf{C}$.

Before solving for $\mathbf{O}$ with subproblem (13), it is necessary to introduce the proximal operator of the function $\phi$. Let $\lambda$, $\gamma$, $\eta$, $\mu \in \mathbb{R}^{+}$, $\textit{x}$, $c \in \mathbb{C}$, we define
\begin{equation}
    \begin{array}{cc}
        {\kern -62pt}{\rm{prox}}_{\mu,\phi(\cdot)}(c)  = \arg\min\limits_{\textit{x}} \mu\phi(\textit{x})+\frac{1}{2}\lvert \textit{x}-c \lvert^2 \\
        {\kern 40pt}\overset{\text{(a)}}{=} \arg\min\limits_{\textit{x}} \mu w^\ast\log\left(\frac{\lvert \textit{x} \lvert}{\eta}+1 \right) 
         +\frac{1}{2} \lvert \textit{x}-c \lvert^2 \\
        {\kern -48pt} \overset{\text{(b)}}{=} {\rm{prox}}_{\mu w^\ast,\log(\cdot)}(c),
    \end{array}
\end{equation}
where in (a) and (b) we respectively used Lemma 1 and Lemma 2. Hence, the solution to $\mathbf{O}$-subproblem (13) can be obtained using the proximal operator of $\phi$, and its specific closed-form solution is 
\begin{equation}
    \begin{array}{cc}
         \mathbf{O}_{m,t}^{l+1}=\text{prox}_{\frac{\lambda_2}{\rho} w^\ast,\log(\cdot)}\left( \mathbf{Q}_{m,t}^l \right), m=1,...,M, t=1,...,T
    \end{array},
\end{equation}
where $\mathbf{Q}^l=\mathbf{Y}-\mathbf{A}\mathbf{X}^{l+1}-\mathbf{W}^l+\frac{\mathbf{U}^l}{\rho}$, $\mathbf{Q}_{m,t}^l$ is the element at the $m$th row and $t$th column of $\mathbf{Q}^l$.

For $\mathbf{W}$-subproblem (14), by equating the derivative of the objective function with respect to $\mathbf{W}$ to zero, we have
\begin{equation}
    \begin{array}{cc}
         \mathbf{W}^{l+1} = -\frac{\rho}{\rho+1} \left(\mathbf{A}\mathbf{X}^{l+1}+\mathbf{O}^{l+1}-\mathbf{Y}-\frac{\mathbf{U}^l}{\rho} \right)
    \end{array}.
\end{equation}

Our proposed algorithm can be summarized as follows:
\begin{algorithm}[H]
    \caption{ADMM Method for Solving $on-grid$ Model}
    \label{alg1}
    \begin{algorithmic}[1]
        \REQUIRE $\mathbf{Y} \in \mathbb{C}^{M\times T},\mathbf{A} \in \mathbb{C}^{N\times T},\lambda_1,\lambda_2 \in \mathbb{R}^+$.
        
       \STATE Initialize $\mathbf{X}^0 = \mathbf{A}^H\mathbf{Y},\mathbf{O}^0 = \mathbf{Y},\mathbf{W}^0 =\mathbf{0}_{M\times T},\mathbf{U}^0 =\mathbf{0}_{M\times T},\lambda=1,\eta=0.5,\gamma=2,\rho=3,\beta=0.03$;
       \STATE Repeat \\
       \quad a) Update $\mathbf{X}^{l+1}$ by (17);\\
       \quad b) Update $\mathbf{O}^{l+1}$ by (20);\\
       \quad c) Update $\mathbf{W}^{l+1}$ by (21);\\
       \quad d) Update $\mathbf{U}^{l+1}$ by (15);\\
       \STATE Until $\Vert\mathbf{X}^{l+1}-\mathbf{X}^l\Vert_F / \Vert \mathbf{X}^l \Vert_F<\epsilon$, where $\epsilon$ is a small threshold (e.g.,$\epsilon=0.0001$).
       \ENSURE $\mathbf{X},\mathbf{O}$.
    \end{algorithmic}
\end{algorithm}

\subsection{Off-Grid Gap Estimation}
As mentioned in Section \uppercase\expandafter{\romannumeral1}, the true DOAs may not be on the predefined grid. To address this issue, we employ the first-order Taylor expansion of the steering matrix. Before that, we need to obtain the on-grid DOAs. We calculate the $l_2$-norm for each row of $\mathbf{X}$ and use the row indices corresponding to the $K$ largest values as on-grid DOAs $\tilde{\boldsymbol{\theta}}=[\tilde\theta_1,...,\tilde\theta_K]^T$.

Specifically, (9) can be rewritten as
\begin{equation}
    \begin{array}{cc}
         \mathbf{Y}=\left(\mathbf{A}(\tilde{\boldsymbol{\theta}}) + \mathbf{B}(\tilde{\boldsymbol{\theta}})\text{diag}(\boldsymbol{\delta})\right)\mathbf{X}+\mathbf{W}+\tilde{\mathbf{O}}
    \end{array},
\end{equation}
where $\mathbf{B}(\tilde{\boldsymbol{\theta}})=\left [\frac{\partial\mathbf{a}(\tilde\theta_1)}{\partial\tilde\theta_1},...,\frac{\partial\mathbf{a}(\tilde\theta_K)}{\partial\tilde\theta_K}  \right]$ is the gradient of $\mathbf{A}(\tilde{\boldsymbol{\theta}})$ with respect to $\tilde{\boldsymbol{\theta}}$, $\boldsymbol{\delta}=\boldsymbol{\theta}-\tilde{\boldsymbol{\theta}} \in \mathbb{R}^{K}$ is the off-grid gap, $\tilde{\mathbf{O}}$ is the estimate at \textbf{Algorithm 1}. The diag$(\cdot)$ function can be used to extract the diagonal elements of a matrix or transform a vector into a diagonal matrix, depending on the context. To obtain the off-grid gap, we formulate the following optimization problem:
\begin{equation}
    \begin{array}{cc}
         {\kern -10pt}\boldsymbol{\delta}=\arg\min\limits_{\boldsymbol{\delta}} \left \Vert \mathbf{Y}-\tilde{\mathbf{O}}-\left(\mathbf{A}(\tilde {\boldsymbol{\theta}}) + \mathbf{B}(\tilde{\boldsymbol{\theta}})\text{diag}(\boldsymbol{\delta}) \right)\mathbf{X} \right \Vert_F^2
    \end{array}.
\end{equation}

Update $\mathbf{X}$ by $\mathbf{\widehat X}=\mathbf{A}^\dagger( \widehat {\boldsymbol{\theta}})(\mathbf{Y}-\tilde{\mathbf{O}})$, where $\widehat {\boldsymbol{\theta}}$ is the estimate of $\boldsymbol{\theta}$. Then, (23) can be expressed as
\begin{equation}
    \begin{array}{cc}
         {\kern -60pt}\boldsymbol{\delta}=\arg\min\limits_{\boldsymbol{\delta}}  \left \Vert \mathbf{H}- \mathbf{B}(\tilde{\boldsymbol{\theta}})\text{diag}(\boldsymbol{\delta}) \mathbf{\widehat X} \right \Vert_F^2 \\
         {\kern 30pt}- 2\mathcal{R}\left(\text{tr}(\mathbf{H}^H \mathbf{B}(\tilde{\boldsymbol{\theta}}) \text{diag}(\boldsymbol{\delta})  \mathbf{\widehat X}    )\right) \\
          = \arg\min\limits_{\boldsymbol{\delta}} \text{tr}(\mathbf{\widehat X}^H \text{diag}(\boldsymbol{\delta}) \mathbf{B}^H (\tilde{\boldsymbol{\theta}}) \mathbf{B}(\tilde{\boldsymbol{\theta}}) \text{diag}(\boldsymbol{\delta})  \mathbf{\widehat X}) \\ 
          {\kern 30pt}- 2\mathcal{R}\left(\text{tr}(\mathbf{H}^H \mathbf{B}(\tilde{\boldsymbol{\theta}}) \text{diag}(\boldsymbol{\delta})  \mathbf{\widehat X}    )\right) \\
         {\kern -18pt}= \arg\min\limits_{\boldsymbol{\delta}} \boldsymbol{\delta}^T((\mathbf{B}^H (\tilde{\boldsymbol{\theta}}) \mathbf{B}(\tilde{\boldsymbol{\theta}})) \odot (\mathbf{\widehat X}\mathbf{\widehat X}^H)^T)\boldsymbol{\delta} \\
         {\kern 20pt}- 2\mathcal{R}\left(\text{diag}(\mathbf{\widehat X}\mathbf{H}^H \mathbf{B}(\tilde{\boldsymbol{\theta}}))^T\boldsymbol{\delta}    \right)
    \end{array},
\end{equation}
where $\mathbf{H}=\mathbf{Y}-\tilde{\mathbf{O}}-\mathbf{A}(\tilde {\boldsymbol{\theta}})\mathbf{\widehat X}, \odot$ is the dot product for a matrix.

By equating the derivative of the objective function with respect to $\boldsymbol{\delta}$ to zero, we obtain
\begin{equation}
    \boldsymbol{\delta}=\mathcal{R}(\mathbf{G}^{-1}\mathbf{z}),
\end{equation}
where $\mathbf{G}=(\mathbf{B}^H (\tilde{\boldsymbol{\theta}}) \mathbf{B}(\tilde{\boldsymbol{\theta}})) \odot (\mathbf{\widehat X}\mathbf{\widehat X}^H)^T)$, and $\mathbf{z}=\text{diag}(\mathbf{\widehat X}\mathbf{H}^H \mathbf{B}(\tilde{\boldsymbol{\theta}}))$. Next, we summarize the proposed off-grid gap estimation algorithm.
\begin{algorithm}[H]
    \caption{$off-grid$ Gap Estimation}
    \label{alg2}
    \begin{algorithmic}[1]
        \REQUIRE $\mathbf{Y},\tilde{\mathbf{O}} \in \mathbb{C}^{M\times T}, \boldsymbol{\theta}^0 \in \mathbb{R}^{K}$.
        
       \STATE Initialize $\boldsymbol{\delta}^0=\mathbf{0}_{K\times1}$;
       \STATE Repeat \\
       \quad a) Update $\mathbf{X}^{l+1}=\mathbf{A}^\dagger(  \boldsymbol{\theta}^l)(\mathbf{Y}-\tilde{\mathbf{O}})$;\\
       \quad b) Update $\boldsymbol{\delta}^{l+1}$ by (25);\\
       \quad c) Update $\boldsymbol{\theta}^{l+1}=\boldsymbol{\theta}^{l}+\boldsymbol{\delta}^{l+1}$;
       \STATE Until $\Vert\boldsymbol{\delta}^{l+1}-\boldsymbol{\delta}^l\Vert_2 / \Vert \boldsymbol{\delta}^l \Vert_2<\epsilon$, where $\epsilon$ is a small threshold (e.g.,$\epsilon=0.0001$).
       \ENSURE $\tilde{\boldsymbol{\theta}}$.
    \end{algorithmic}
\end{algorithm}

\section{Simulation Results}
In this section, we present several examples of the proposed algorithm. We consider a ULA with 10 elements. Two narrowband sources with equal power impinge on the ULA. Specifically, the uncorrelated source waveforms are generated as $s_k=e^{j\varphi(s_k)}$, and the phase $\varphi(s_k)$ follows a uniform distribution in the range $[0,2\pi]$. Unless otherwise specified, these two sources come uniformly from the interval $[-10^\circ,0^\circ]$ and $[20^\circ,30^\circ]$. $\mathbf{W}$ is an independent and identically distributed complex Gaussian noise matrix with mean zero and variance of $\sigma_1^2$. $\mathbf{O}=\mathbf{E}\odot \mathbf{N}$ is used to model outliers, where $\mathbf{E}$ is Bernoulli distributed with $P\left\{ \mathbf{E}_{m,t}=1 \right\} =p$, $p$ is the probability of outliers occurring, and $\mathbf{N}$ is a complex Gaussian noise matrix with mean zero and variance  $\sigma_2^2$. The signal-to-noise ratio (SNR) and the signal-to-outlier ratio (SOR) are defined by $\sigma_s^{2}/\sigma_1^{2}$ and $\sigma_s^{2}/\sigma_2^{2}$, respectively, where $\sigma_s^{2}$ is variance of the source. In each simulation, 1000 Monte Carlo experiments are conducted and SOR = -20dB. The average RMSE is given by
\begin{equation}
    \begin{aligned}
        \text{RMSE}=\sqrt{\frac{\sum_{j=1}^{N_{mc}}\sum_{k=1}^{K}(\tilde{\theta}_{j,k}-{\theta}_k)}{N_{mc}K}},
    \end{aligned} 
\end{equation}
where $N_{mc}$ is the number of Monte Carlo experiments, $\tilde{\theta}_{j,k}$ is the value estimated for the $k$th source in the $j$th Monte Carlo experiment, and ${\theta}_k$ is true DOA for the $k$th source. Given that some algorithms may not perform well in extreme cases, leading to a very large RMSE, we include cases where the DOA estimation error is less than $3^\circ$ in the RMSE. $l_p$-MUSIC \cite{zeng2013ell}, Bayes-optimal \cite{dai2017sparse}, SOMP-LS \cite{gretsistas2012alternating}, GMCCC-MUSIC \cite{ma2020generalised}, and SBL-Tyler \cite{mecklenbrauker2023robust} are used for comparison. The grid interval is selected as $r=2^\circ$, and $p=1.1$ for $l_p$-MUSIC. For SBL-Tyler, as recommended, the dictionary size $N=18001$. We tested the relationship between SNR and DOA estimation accuracy. According to the results, when the SNR is greater than $5$dB, we use $\lambda_1=7,\lambda_2=1.4$; otherwise, we use $\lambda_1=4,\lambda_2=4$. 

In the first example, the performance of various algorithms is tested under different SNRs. The number of snapshots is $T$ = 30 and $p$ is set to 0.1. From the left side of Fig. 2, it can be observed that SOMP-LS exhibits the poorest performance, mainly due to its high sensitivity to outliers. As the SNR increases, $l_p$-MUSIC and GMCCC-MUSIC show a reduced rate of performance improvement due to the limitation of snapshots. In contrast, methods based on sparse representation do not have strict requirements in this regard. Compared to Bayes-optimal, our proposed algorithm can more accurately mitigate the influence of outliers, leading to improved performance.                                  

\begin{figure}[ht]
    \centering
    \includegraphics[scale=0.19]{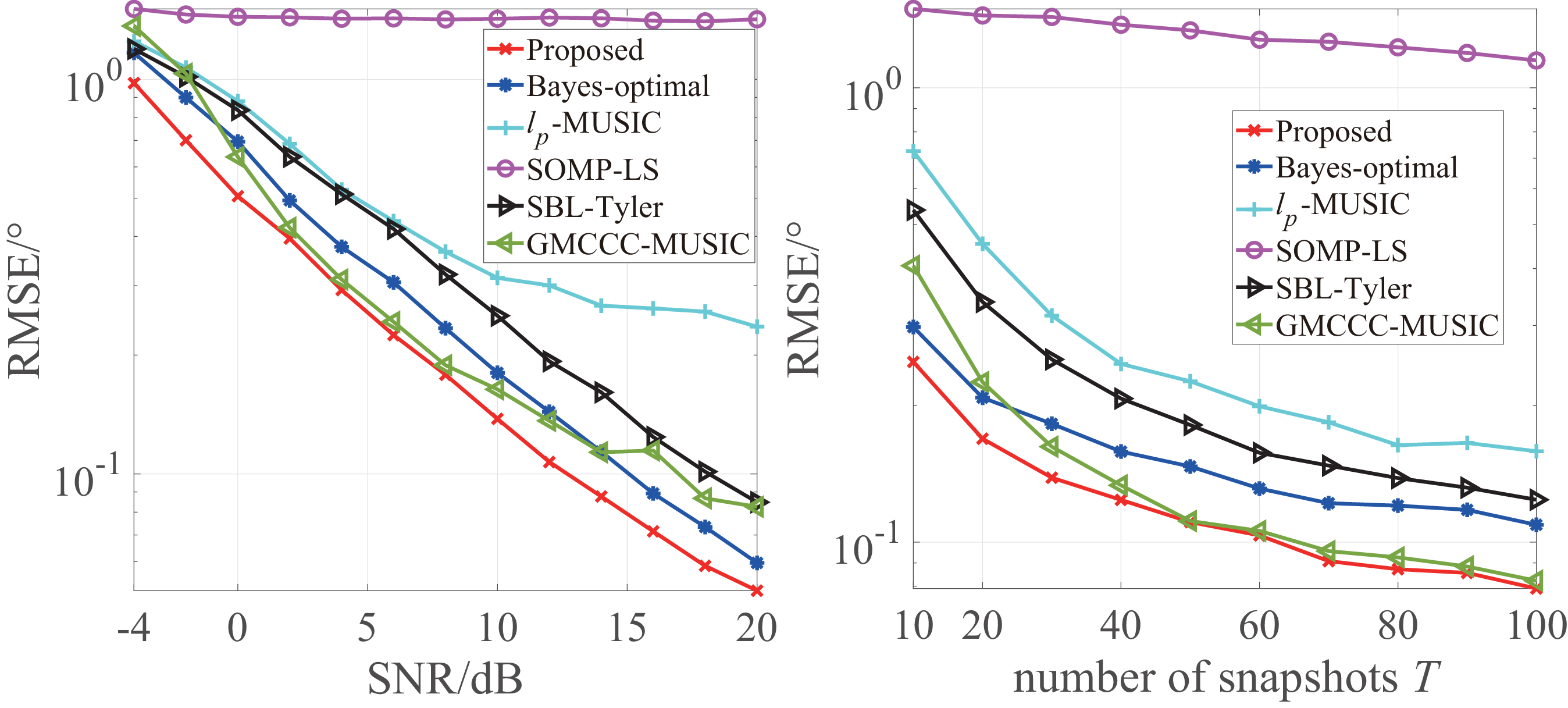}
    \caption{\small RMSE of uncorrelated DOA estimates for (left) different SNR and for (right) different number of snapshots, with $M=10$ and $p=0.1$.}
    \label{SNR_T}
\end{figure}

In the second example, we investigate the impact of the number of snapshots on DOA estimation performance. The SNR and $p$ are fixed at 10dB and 0.1 respectively. As demonstrated in the right side of Fig. 2, the proposed algorithm exhibits greater robustness. Interestingly, as the number of snapshots increases, GMCCC-MUSIC outperforms Bayes-optimal. This is because subspace-based algorithms heavily rely on large samples.

\begin{figure}[ht]
    \centering
    \includegraphics[scale=0.19]{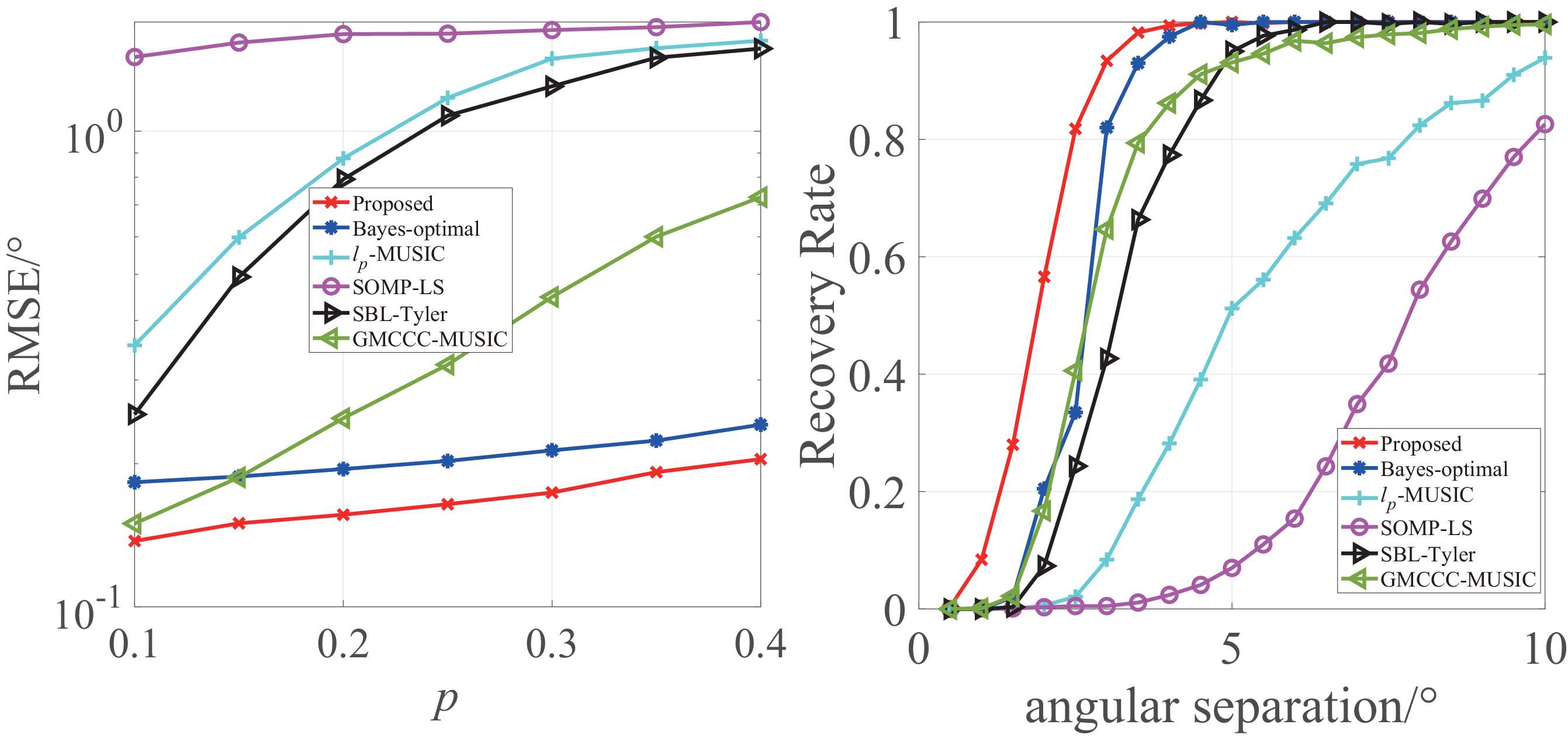}
    \caption{\small RMSE of uncorrelated DOA estimates for (left) different outlier probability ($p$) and (right) for different angular separation, with SNR = 10dB and $T$ = 30.}
    \label{T}
\end{figure}

The left side of Fig. 3 presents the variation in algorithm performance with changes in the probability of outlier occurrences. The proposed algorithm and Bayes-optimal demonstrate stable performance, which is attributed to the separate handling of outliers in the presence of impulse noise.

In DOA estimation, the ability to distinguish between two closely located sources is one of the important criteria for assessing performance. In the fourth example, $p$ are set to 0.1. The two uncorrelated sources are considered with DOAs $\theta_1=-10.8^\circ$ and $\theta_2=-10.8^\circ+\Delta\theta$, where $\Delta\theta$ varies from $0.5^\circ$ to $10^\circ$. We consider the two sources  distinguishable if $\max\limits_{k=1,2} \lvert \tilde{\theta}_k - \theta_k \lvert$ is less than $\lvert \theta_1 - \theta_2 \lvert/2$. From the right side of Fig. 3, we notice that our algorithm outperforms other methods.

In the last example, the performance of the algorithms under coherent (fully correlated) sources is investigated, where the parameters are the same as in example 1. From Fig. 4, we observe that the proposed algorithm has good performance in this environment, while the performance of the $l_p$-MUSIC and GMCCC-MUSIC algorithms significantly degrades.

\begin{figure}[ht]
    \centering
    \includegraphics[scale=0.35]{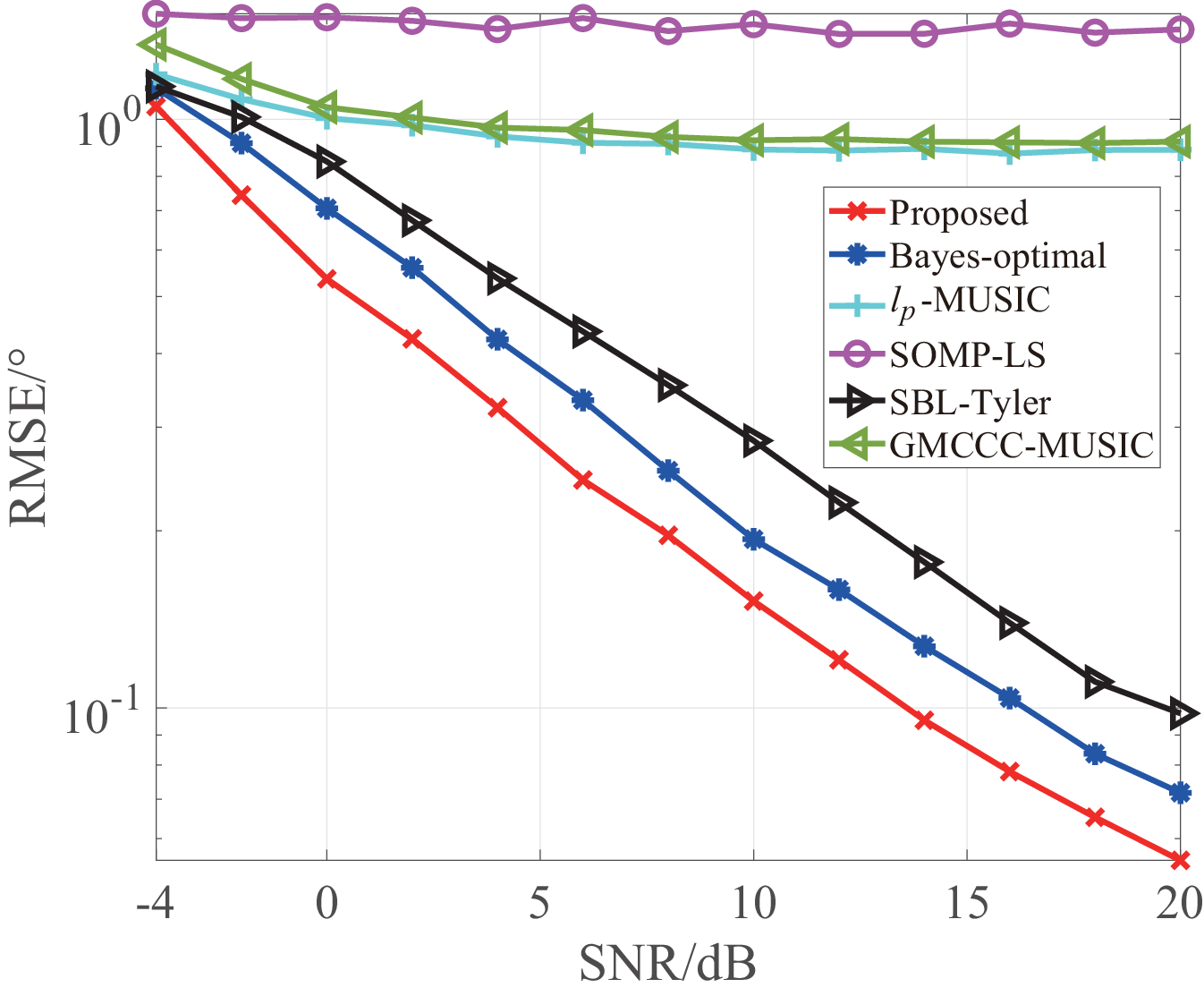}
    \caption{\small RMSE of coherent DOA estimates for different SNR.}
    \label{COH}
\end{figure}

\section{Conclusions}

We have introduced a novel DOA estimation algorithm with robustness against impulsive noise. We model impulsive noise as a combination of outliers and Gaussian noise, exploiting the statistical differences between them. By employing the ADM to solve the optimization problem, we obtain outliers matrix and the on-grid DOAs. To address the grid mismatch, we use an alternating approach with the obtained outlier matrix and on-grid DOAs to obtain the final DOAs. Simulation results demonstrate that this method exhibits excellent resilience against impulsive noise.

\bibliographystyle{IEEEtran}
\vfill
\bibliography{IEEEabrv, myref} 

\end{document}